# Large Language Model Consensus and Corpus-Adaptive Topic Modeling for Classifying and Mapping Patent Innovation

**Sercan Ozcan [1,2,3,*], Enes Ozgozler [4], C. Okan Sakar [4]**

[1] School of Strategy, Marketing and Innovation, University of Portsmouth, UK

[2] Innovative Management Center, Azerbaijan State University of Economics (UNEC), Azerbaijan

[3] Faculty of Economics and Administrative Sciences, Piri Reis University, Turkey

[4] Department of Computer Engineering, Bahcesehir University, Istanbul, Turkey

*Corresponding author: sercan.ozcan@port.ac.uk

**Abstract**

Patent analytics increasingly relies on transformer representations and large language models (LLMs), yet linking conceptual classification, topic formation and interpretation requires explicit decisions about model roles and the evidence retained at each stage. In this design-oriented study, we propose and demonstrate a patent analytics pipeline that links these tasks and keeps each decision open to audit. Three LLMs separately classify each patent by novelty (radical or incremental) and innovation type (product, process, service). Radical and innovation-type labels each require two supporting votes, with Incremental used as the fallback novelty label. Topics are discovered by clustering PatentSBERTa embeddings, with parameter ranges scaled to corpus size. A patent-specific stopword list is applied only when topics are represented with class-based term frequency–inverse document frequency (c-TF-IDF), and a second embedding model supports inter-topic similarity, hierarchical refinement and visualization. An LLM then labels each topic from its ranked keywords and representative patents. We demonstrate the pipeline on 30,964 public-sector-related patents retrieved from Lens.org. Across five category corpora ranging from 591 to 29,215 patents, the pipeline produced 74 category-specific topic instances with a macro-average document coverage of 98.1% and topic diversity of 0.900–1.000, and fewer than 1% of classification outputs returned UNKNOWN. Because every model vote and the evidence behind each topic label are retained, analysts can audit and revise the outputs. The demonstration establishes the feasibility and traceability of the pipeline; testing its labels against expert coding and benchmarking it against alternative pipelines are identified as the next steps. The pipeline offers a reusable basis for LLM-assisted patent analysis in technology intelligence.

**Keywords:** patent analysis; large language models; radical innovation; innovation type; topic modeling; technology intelligence

## 1. Introduction

Patent documents are a central source of evidence for technology intelligence because they combine technical descriptions with bibliographic, legal, and classification information at a scale that is difficult to reproduce through other innovation data sources. Patent analysis has therefore been used to map technological knowledge, monitor emerging fields, compare knowledge profiles, forecast technological trajectories, and support technology roadmapping. The difficulty is that the most informative parts of a patent are textual, terminologically specialized, and semantically heterogeneous. As patent corpora grow, keyword counts, citation indicators, and manually defined taxonomies are no longer sufficient on their own to identify technological structure.

Computational methods for patent analysis have developed in several stages. Earlier work used keyword extraction, latent topic models, and unsupervised learning to identify technological domains and emerging patterns (Joung and Kim, 2017; Chen et al., 2017; Suominen et al., 2017). Later studies combined patent indicators with supervised machine learning and deep learning for emerging-technology detection, technological forecasting, identification of promising technologies, and patent-value analysis (Kyebambe et al., 2017; Lee et al., 2018; Zhou et al., 2020; Choi et al., 2021; Trappey et al., 2021). More recent work has used distributional semantics and transformer models such as BERT to measure patent similarity and identify emerging topics (Song et al., 2023; Bekamiri et al., 2024; Rezende et al., 2024). This work has moved patent analysis from lexical matching toward representations of technical meaning.

Topic modeling has changed in parallel. LDA remains common in patent roadmapping (Zhang et al., 2021), but embedding-based approaches encode documents in a dense semantic space, reduce its dimensionality, cluster the documents by density, and describe each cluster with class-specific term weights. BERTopic implements this approach with sentence embeddings, UMAP, HDBSCAN, and c-TF-IDF (Grootendorst, 2022). BERTopic has already been applied to patents, for example, with PatentSBERTa embeddings (Jeon et al., 2023) and in combination with S-curve analysis for roadmapping (Maghsoudi et al., 2025). Applied to patents, however, a default implementation leaves several choices open: which embedding model to use, how to treat legal boilerplate, how to set parameters for corpora of different sizes, how to handle outliers and near-duplicate topics, and how to produce readable topic labels.

LLMs open a second line of development. Given task definitions, examples, and an output format, generative models can annotate text and justify their labels at a scale that manual coding cannot match (Gilardi et al., 2023; Ziems et al., 2024). Pelaez et al. (2024), for example, used GPT-4 to label public value expressions in AI patents, and Tan et al. (2024) stress that LLM annotations must be assessed before they are used. Relying on a single model, however, ties the results to that model's interpretations, prompt sensitivity, and failure modes. Treating the outputs of several models as separate judgments makes these dependencies visible, although different models can share errors (Kim et al., 2025).

The individual techniques are well established. An integrated patent workflow must additionally specify how conceptual labels define the corpora to be clustered and how later outputs can be traced to earlier decisions. These links matter because classification errors and refinement choices can alter the technological map that analysts ultimately interpret. We address this gap with four design principles: (i) consensus-based conceptual classification that makes model agreement and disagreement observable; (ii) representation-task separation, whereby document embedding and lexical topic representation use task-appropriate preprocessing; (iii) corpus-sensitive topic formation and refinement instead of applying a fixed clustering configuration across corpora; and (iv) evidence-grounded interpretation that preserves traceability from topic labels to statistical topic representations and representative patents. Our contribution lies in how these components are combined, not in the components themselves.

We design such a pipeline for innovation classification and topic discovery and demonstrate it on a corpus of public-sector-related patents. Drawing on the design science perspective of Hevner et al. (2004), this study specifies an analytical pipeline and demonstrates its application. We address three design-oriented research questions:

- **RQ1.** How can several LLMs be combined to classify patents on two distinct innovation dimensions, novelty and innovation type, while keeping each model's decision open to inspection?
- **RQ2.** How can embedding-based topic modeling be configured so that a single procedure yields usable topic structures for patent corpora that differ in size by more than an order of magnitude?
- **RQ3.** How can LLM-generated topic labels be kept traceable to the statistical and documentary evidence from which they are derived?

The study makes four contributions. First, it specifies a dimension-wise multi-LLM consensus architecture in which novelty and innovation type are classified separately, allowing multi-label type assignments while retaining an explicit record of model agreement and disagreement. Second, it builds a patent-specific stopword list from a multi-domain patent sample and applies it only at the c-TF-IDF stage, leaving the abstracts unaltered for embedding. Third, it specifies how UMAP, HDBSCAN, and vectorizer settings scale with corpus size, together with rules for outlier reassignment and hierarchical merging. Fourth, it labels topics from both ranked keywords and representative patents, so that each label can be traced to its evidence; PatentSBERTa is used for document clustering and Qwen3-Embedding for topic-level similarity. Together, these elements form a reusable pipeline for patent-based technology intelligence.

The remainder of this paper is organized as follows. Section 2 reviews related work and derives the design principles of the pipeline. Section 3 describes the pipeline, and Section 4 sets out the demonstration design. Section 5 presents the results and answers the research questions. Section 6 discusses the contribution of the study and its implications for research and practice, and Section 7 concludes with the limitations of the study and directions for future research.

# 2. Related Work and Methodological Positioning

## 2.1 From patent indicators to semantic technology intelligence

Patent analytics has long combined bibliographic indicators, citation structures, and text to infer technological activity. Unsupervised learning has been used to map firms' knowledge profiles (Suominen et al., 2017), while topic-based approaches have supported technological forecasting (Chen et al., 2017) and technology planning through technical keywords (Joung and Kim, 2017). Patent analytics has also been integrated into roadmapping, including LDA-based technology assessment in blockchain (Zhang et al., 2021), and into trend forecasting and technology opportunity discovery that combine patents with scientific publications (Wang et al., 2022). This stream treats patent text as a source of latent technological structure, not only as a count of applications or citations.

Supervised and deep-learning approaches extend this tradition by treating patent analytics as a prediction and classification problem. Early identification studies combine multiple patent indicators with machine learning (Lee et al., 2018), while later work applies deep learning to patent value, emerging technology identification and risk forecasting (Zhou et al., 2020; Trappey et al., 2021; Chi and Wang, 2022). These studies show the value of computational modeling, but they predict a predefined output such as an emerging-technology label, value score or risk class. Topic discovery differs because the relevant technological categories are not known in advance.

## 2.2 Semantic representation, transformer models and patent topic discovery

Semantic patent analysis attempts to represent technological meaning beyond exact vocabulary overlap. Technological distance has been measured for patent mapping (Yan and Luo, 2017), semantic and citation information have been combined for technology forecasting (Qiu and Wang, 2022), and NLP approaches have compared LSA, word2vec, and Word Mover's Distance for patent similarity and forecasting (Rezende et al., 2024). Sentence-BERT established an efficient semantic-similarity architecture based on siamese transformer encoders (Reimers and Gurevych, 2019), and patent-specific transformer applications include BERT-based identification of emerging technology topics (Song et al., 2023) and domain-specific sentence encoders such as PatentSBERTa (Bekamiri et al., 2024). PatentSBERTa is relevant here because it was fine-tuned on patent claims to measure technological similarity between patents.

Embedding-based topic modeling separates the semantic representation of documents from the term-frequency representation of topics. BERTopic combines dense document representations with UMAP, HDBSCAN and c-TF-IDF (Grootendorst, 2022). UMAP provides nonlinear dimensionality reduction (McInnes et al., 2018), while HDBSCAN identifies density-based clusters and can treat low-density documents as noise (Campello et al., 2013). This architecture suits patent corpora because related inventions described in different technical terms can still lie close together in the embedding space. The resulting topics, however, depend on the embedding model, the density parameters, the treatment of outliers and the handling of frequent patent-specific terms.

Recent patent studies combine several analytical layers: BERTopic with S-curve analysis for circular-economy roadmapping (Maghsoudi et al., 2025), structural topic modeling for roadmapping applications of generative AI (Singh et al., 2024), topic modeling and trend analysis of AI patents for climate change (Podrecca et al., 2024), and CPC-based diffusion modeling and clustering for monitoring technological innovation (Lee et al., 2026). As such pipelines become common, a new pipeline must show what its particular combination of components adds.

## 2.3 LLM-based annotation and consensus classification

Generative language models are useful for patent classification when the categories of interest are conceptual and not encoded in patent office taxonomies. Brown et al. (2020) showed that large generative models can perform new tasks from a few examples, and later work found that prompting models to state intermediate reasoning steps improves performance on complex tasks (Wei et al., 2022). In patent research, Pelaez et al. (2024) used a generative language model to produce labels and rationales for large-scale analysis of public value expressions in AI patents. They showed that conceptual patent labels can be generated at scale when the task is carefully specified; agreement with GPT-4's labels reached up to 95% on a separate 1,000-sentence evaluation set.

Automated annotation, however, raises a reliability problem. A single LLM may be internally consistent yet systematically wrong, sensitive to wording, or biased toward particular category interpretations. Tan et al. (2024) accordingly describe LLM annotation as a sequence of generation, assessment, and use, in which assessment cannot be skipped. We respond by obtaining separate judgments from three different models and applying a majority rule for each dimension, an approach related to panels of LLM judges (Verga et al., 2024). Consensus is no substitute for a human gold standard, and errors can be correlated across models (Kim et al., 2025), but voting reduces dependence on any one model and records where the models disagree.

## 2.4 Conceptual dimensions of innovation classification

Innovation classification requires explicit definitions of both the degree of novelty and the form of innovation. Garcia and Calantone (2002) document inconsistencies in innovation terminology and argue that innovativeness should be assessed from both marketing and technological perspectives, at the macro and micro levels. The radical/incremental distinction concerns the extent of departure from established approaches, whereas innovation type concerns the object of change. The Oslo Manual distinguishes product innovation, encompassing goods and services, from business process innovation, which concerns production, delivery and other business functions; these types can occur in combination (OECD and Eurostat, 2018). Novelty and innovation type therefore provide conceptually distinct dimensions for characterizing an invention.

Patent research also distinguishes the assessment of technological novelty from the observation of subsequent influence. Dahlin and Behrens (2005) define technological radicalness through novelty, uniqueness and impact on future technology, and measure it through overlaps in patent citations. Later measures capture novelty as new combinations of technology classes (Verhoeven et al., 2016) or as textual dissimilarity from earlier patents (Arts et al., 2021; Kelly et al., 2021). Dahlin and Behrens' framework separates the identification of potentially radical inventions from the retrospective assessment of their technological impact. This distinction is relevant to text-based classification: patent descriptions provide evidence about the proposed technical contribution, but do not by themselves establish subsequent adoption or technological influence. Accordingly, a novelty label inferred from patent text requires a stated frame of comparison and should not be interpreted as a direct measure of realized impact.

In the present study, innovation classification complements topic discovery by characterizing the novelty and form of inventions within the technological themes identified in the corpus. The radical/incremental distinction concerns departure from established technological approaches in the relevant field. Product, Process and Service are retained as separate analytical labels so that service-oriented inventions can be examined explicitly. This choice is an adaptation for patent analysis rather than a direct reproduction of the Oslo Manual taxonomy, in which services are included within products (OECD and Eurostat, 2018). The labels may overlap when an invention combines different forms of innovation. Section 3.3 specifies the label structure and consensus procedure used to operationalize these distinctions. These labels describe interpretations of disclosed inventions; they do not establish implementation or realized innovation.

### 2.5 Synthesis: Four design principles for an integrated framework

From these strands of work, we derive four design principles for a patent analytics pipeline. First, conceptual classification should be auditable rather than dependent on one generative model. Second, semantic document representation and lexical topic representation should be treated as different tasks. Third, density-based topic formation and refinement should be adapted to corpus scale rather than rely on a universal parameterization. Fourth, machine-generated topic interpretation should remain anchored to statistical topic evidence and representative documents. Existing studies supply the individual components; our aim is to combine them in one traceable workflow. Table 1 summarizes this positioning.

**Table 1.** Methodological streams and positioning of the proposed framework.

| Methodological stream | Representative studies | Established capability | Design principle addressed here |
|---|---|---|---|
| Patent mapping and technology intelligence | Suominen et al. (2017); Chen et al. (2017); Joung and Kim (2017); Zhang et al. (2021) | Mapping, forecasting and roadmapping from patent data | Evidence-grounded, traceable outputs for downstream technology intelligence |
| Semantic patent representation | Yan and Luo (2017); Qiu and Wang (2022); Rezende et al. (2024); Bekamiri et al. (2024) | Technological distance and semantic similarity | Representation-task separation through patent-domain document encoding |
| Embedding-based topic modeling | McInnes et al. (2018); Campello et al. (2013); Grootendorst (2022); Song et al. (2023) | Semantic clustering and topic representation | Corpus-sensitive topic formation and refinement |
| LLM annotation | Brown et al. (2020); Wei et al. (2022); Pelaez et al. (2024); Tan et al. (2024) | Flexible conceptual text labeling and structured outputs | Consensus-based conceptual classification and auditability |
| Integrated patent analytics | Singh et al. (2024); Podrecca et al. (2024); Maghsoudi et al. (2025); Lee et al. (2026) | Multi-stage technology monitoring and roadmapping | Coordination of all four design principles in one architecture |

## 3. Methodology: Integrated Patent Analytics Framework

### 3.1 Framework architecture

The four design principles are implemented across four operational layers: corpus construction, innovation classification, corpus-adaptive topic discovery, and interpretation. Figure 1 separates these layers explicitly. The principles explain why tasks are separated; the layers show where each task is carried out. Domain filtering is optional; innovation classification is conceptually distinct from unsupervised topic formation; topic representation is separated from document embedding; and interpretation occurs only after the topic structure has been formed and refined. No single model or representation is responsible for the whole analysis.

Algorithm 1 gives the complete procedure. Given a collection of patent records, the procedure produces document-level innovation classifications, category-specific topic structures, inter-topic similarity estimates, and evidence-grounded topic labels. Model-level decisions and intermediate analytical outputs are retained to support traceability and reproducibility.

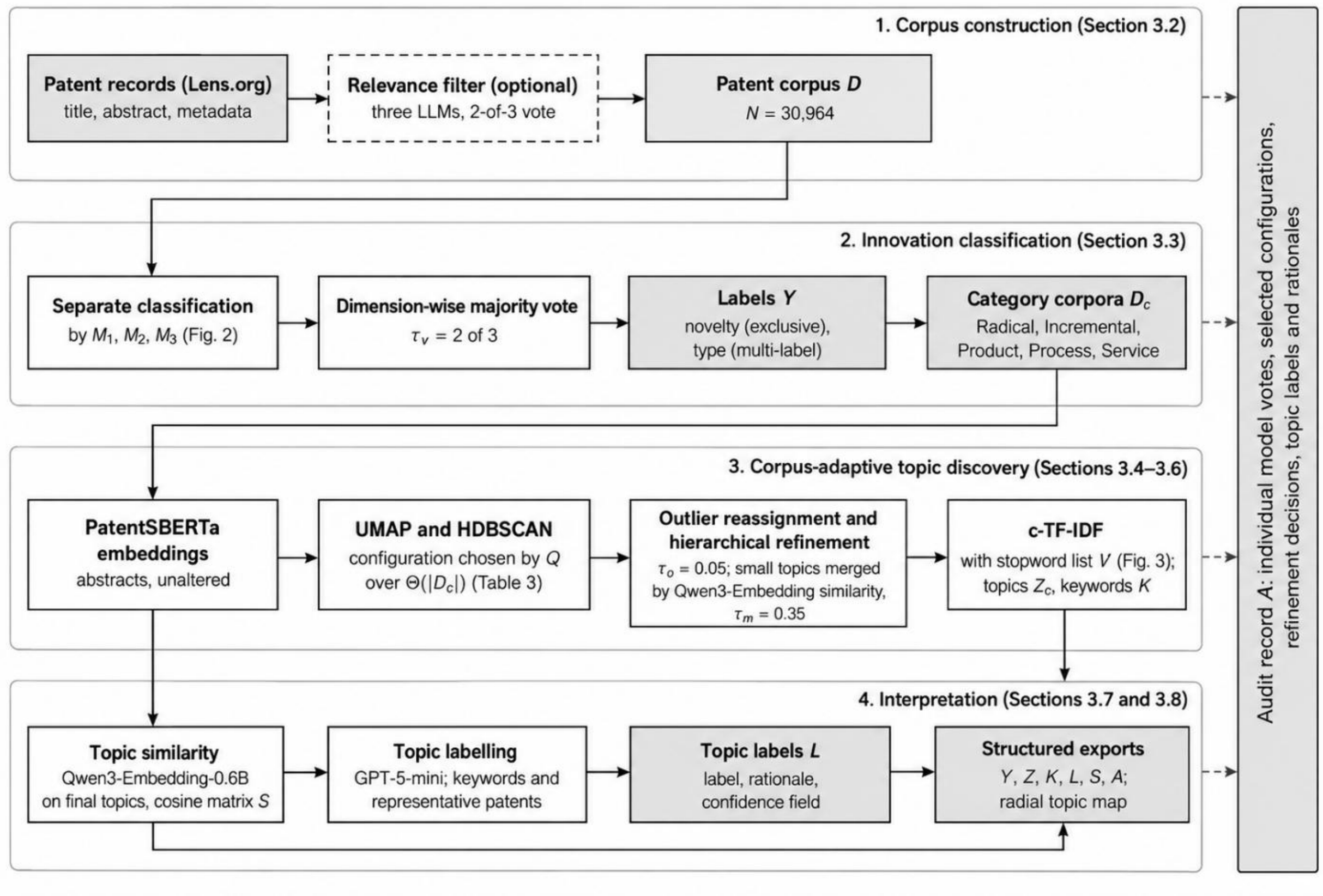

**Figure 1.** Integrated AI-based patent analytics framework.

## 3.2 Corpus acquisition and optional domain-relevance filtering

Patent documents were retrieved from Lens.org through a search conducted on 26 April 2025. The search combined 38 terms referring to local-government institutions, administration, infrastructure, and services across titles, abstracts, and claims, with OR applied within and across fields. Retrieval was restricted to records classified by Lens as active and published from 1 January 2014 onward, with no restrictions on jurisdiction or document type. Appendix A provides the complete search strategy and the composition of the initial corpus. After deduplication by Lens ID, the corpus comprised 46,375 patent documents, including 695 (1.5%) without abstracts. The analytical procedure used English-language titles and abstracts. Of these, 41,107 (88.6%) were Chinese publications, 3,181 were from the United States, 647 were from the European Patent Office, and 1,440 were from other jurisdictions; by document type, 34,809 were Chinese utility models (Lens "Limited Patent"), 8,310 were applications, 3,181 were granted patents, and 75 were other documents.

Three language models, Kimi K2, DeepSeek-V3.1, and gpt-oss-120b, independently assessed public-sector relevance. A patent is retained when at least two models classify it as relevant. Each of the three LLMs receives the same document evidence and the same task specification. Every model vote is stored. Innovation classification applies the two-vote threshold with the fallback rule described in Section 3.3. The final analytical corpus contained 30,964 documents (66.8% of the initial corpus). The unit of analysis is the patent document, encompassing applications, granted patents, and utility models. The findings therefore characterize patent-disclosed technologies judged relevant to public-sector functions; they do not establish government ownership, municipal adoption, or realized public value.

### 3.3 Multi-LLM consensus for innovation classification

The classification layer operationalizes the two conceptually distinct dimensions introduced in Section 2.4. Novelty is encoded as a mutually exclusive binary label: Radical or Incremental. Innovation type is encoded as a multi-label output: Product, Process, and Service are evaluated separately, allowing a patent to receive more than one type label. Table 2 summarizes the categories, their analytical roles, and their output structures.

**Table 2.** Operationalization of the two innovation-classification dimensions.

| Dimension | Categories | Role in the analytical pipeline | Output structure |
|---|---|---|---|
| Innovation novelty | Radical; Incremental | Conceptual classification of the degree of technological novelty/departure | Mutually exclusive binary decision |
| Innovation type | Product; Process; Service | Conceptual classification of how an innovation is embodied or delivered | Each type evaluated separately; multi-label outcomes permitted |

Three LLMs, Kimi K2 (kimi-k2:1t; Kimi Team, 2025), DeepSeek-V3.1 (deepseek-v3.1:671b; DeepSeek-AI, 2025), and gpt-oss-120b (gpt-oss:120b; OpenAI, 2025b), are queried separately with the same patent text. All three models were accessed through the Ollama Cloud service (model tags kimi-k2:1t-cloud, deepseek-v3.1:671b-cloud, and gpt-oss:120b-cloud; API-key access) via the Ollama Python client (ollama-python 0.6.0) at temperature 0, and patents were submitted in batches of ten. Each prompt defines the model's role, a decision rule for each category, and a JSON output schema. The final labels follow dimension-wise vote thresholds and the novelty fallback rule, while the individual model decisions are retained for inspection. An output that cannot be parsed after repeated attempts is recorded as UNKNOWN and counts as no vote: a patent is labeled Radical only when two models vote Radical and is otherwise labeled Incremental, and an innovation type is assigned only when two models support it. Consequently, the Incremental category can include fallback assignments without two explicit Incremental votes; these cases can be identified from the stored vote records.

Let $m = 1, 2, 3$ index the three models, and let $v_{im}^{\text{nov}}$ denote the novelty vote of model $m$ for patent $i$. Patent $i$ receives the Radical label when at least $\tau_v = 2$ models vote Radical and is otherwise assigned Incremental:

$$y_i^{\text{nov}} = \begin{cases} \text{Radical}, & \text{if} \sum_m \mathbb{I}(v_{im}^{\text{nov}} = \text{Radical}) \geq \tau_v \\ \text{Incremental}, & \text{otherwise} \end{cases} \quad (1)$$

For each innovation type $t \in \{\text{Product, Process, Service}\}$, let $v_{im}^t = 1$ if model $m$ assigns type $t$ to patent $i$, and $v_{im}^t = 0$ otherwise. Each type is then voted on independently:

$$y_{it} = \begin{cases} 1, & \text{if} \sum_m \mathbb{I}\left(v_{im}^t = 1\right) \geq \tau_v \\ 0, & \text{otherwise} \end{cases}, \quad t \in \{\text{Product, Process, Service}\} \quad (2)$$

The second rule allows a patent to receive more than one type label when it combines several forms of innovation. The consensus architecture is illustrated in Figure 2. Consensus is not proof of validity; its value is that separate votes make agreement and disagreement observable and reduce dependence on any one model.

***Algorithm 1.*** Integrated multi-LLM and corpus-adaptive patent analytics procedure

**Input**

$D = \{d_1, d_2, \dots, d_N\}$: patent records containing titles, abstracts, and available metadata.
$M = \{M_1, M_2, M_3\}$: classification language models.
$H$: heterogeneous patent sample used to construct patent-specific low-information terms.
$\Theta(n)$: corpus-size-dependent UMAP, HDBSCAN, and vectorization configuration space.
$\tau_v = 2$: majority-vote threshold; $\tau_o = 0.05$: outlier-reassignment threshold; $\tau_m$: hierarchical topic-merging threshold.

**Output**

$Y$: document-level novelty and innovation-type labels; $Z$: category-specific topic assignments; $K$: ranked topic keywords and representative patents; $L$: structured topic labels, rationales, and confidence fields; $S$: inter-topic similarity matrices; $A$: audit records containing model votes and intermediate outputs.

**Procedure**

1: Initialize the audit record $A$.
2: **if** domain-relevance filtering is enabled **then**
3: **for each** patent $d_i \in D$ **do**
4: Obtain an independent relevance decision $r_{im}$ from each model $M_m \in M$.
5: Retain $d_i$ if $\sum_m \mathbb{I}\,(r_{im} = \text{relevant}) \geq \tau_v$.
6: Store all model-level relevance decisions in $A$.
7: **end for**
8: Replace $D$ with the retained patent set.
9: **end if**
10: **for each** patent $d_i \in D$ **do**
11: Obtain novelty vote $v_{im}^{\text{nov}} \in \{\text{Radical,Incremental}\}$ from each model $M_m \in M$.
12: Assign the final novelty label using the $\tau_v$-vote threshold and Incremental fallback (Eq. 1).
13: **for each** innovation type $t \in \{\text{Product,Process,Service}\}$ **do**
14: Obtain the binary type decision $v_{im}^{t}$ from each model $M_m \in M$.
15: Set $y_{it} = 1$ if $\sum_m \mathbb{I}\,(v_{im}^{t} = 1) \geq \tau_v$; otherwise set $y_{it} = 0$ (Eq. 2).
16: **end for**
17: Store the individual model votes and final classifications in $A$.
18: **end for**
19: Construct the patent-specific low-information term set $V$ from $H$ (Eq. 3):
$V = V_{\text{baseline}} \cup V_{\text{multi-domain}} \cup (V_{\text{TF-IDF}} \cap V_{\text{BM25}}) \cup V_{\text{patent-jargon}}$.
20: Define the category-specific corpora $D_c$ for $c \in \{\text{Radical,Incremental,Product,Process,Service}\}$ according to the final classifications in $Y$.
21: **for each** category-specific corpus $D_c$ **do**
22: Generate PatentSBERTa embeddings $E_c$ from the patent abstracts without stopword removal.
23: Construct the candidate configuration set $\Theta_c = \Theta(|D_c|)$.
24: **for each** configuration $\theta \in \Theta_c$ **do**
25: Reduce $E_c$ using UMAP under configuration $\theta$.
26: Generate provisional topic assignments $Z_{c,\theta}$ using HDBSCAN.
27: Construct provisional c-TF-IDF topic representations using $V$.
28: Calculate document coverage $C_{c,\theta}$, rescaled silhouette score $\text{Sil}_{c,\theta}$, and topic diversity $\text{Div}_{c,\theta}$.
29: Calculate the configuration score (Eq. 5):
$Q_{c,\theta} = 0.40C_{c,\theta} + 0.30\text{Sil}_{c,\theta} + 0.30\text{Div}_{c,\theta}$.
30: **end for**
31: Select $\theta_c^* = \arg\max_{\theta \in \Theta_c} Q_{c,\theta}$.
32: Retain the topic solution produced under $\theta_c^*$.
33: Reassign eligible HDBSCAN outliers whose assignment probability satisfies the threshold $\tau_o$.
34: Apply hierarchical refinement to small or semantically redundant topics using Qwen3-Embedding topic similarities (Eq. 4) and threshold $\tau_m$.

Recompute the final c-TF-IDF representations and ranked topic keywords.
Store the selected configuration, topic assignments, and refinement decisions in $A$.
**end for**
**for each** final topic $T_k$ **do**
Generate a topic-level embedding using Qwen3-Embedding from the final ranked keywords
Calculate cosine similarities between $T_k$ and the other topics (Eq. 4).
Select the ten highest-ranked c-TF-IDF keywords and ten representative patents for $T_k$.
Provide this evidence to the topic-labeling LLM.
Generate a concise topic label, rationale, and confidence field under the predefined structured-output schema.
**end for**
Export $Y, Z, K, L, S$ and $A$ in structured machine-readable formats.
**return** $Y, Z, K, L, S, A$.

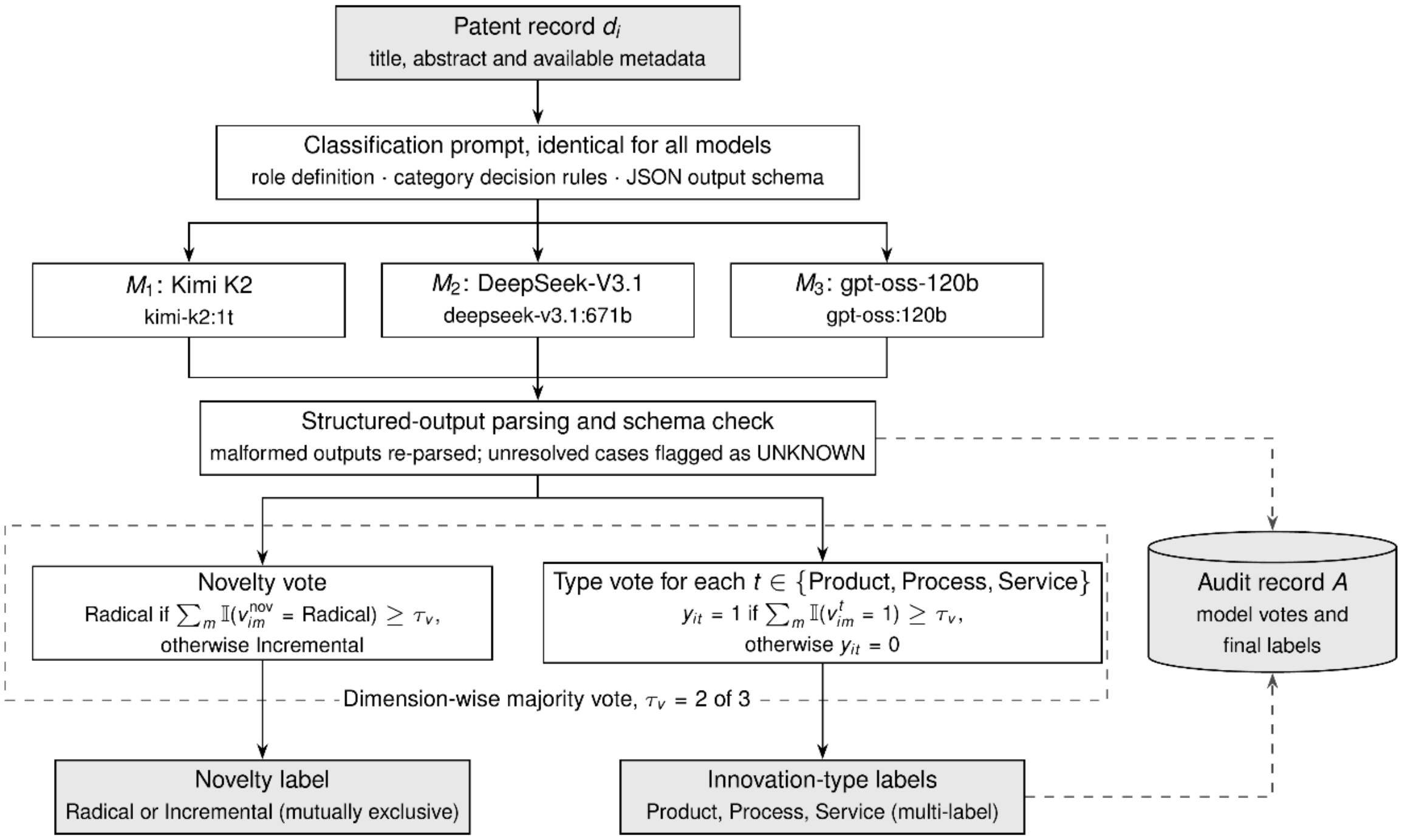

**Figure 2.** Dimension-wise multi-LLM consensus for patent innovation classification.

### 3.4 Patent-specific stopword construction and delayed preprocessing

Semantic and lexical representations are kept separate. Stopwords are not removed before document embedding. Patent abstracts are passed to the embedding model unaltered, so that technical phrases and their context remain available to the transformer. Low-information terms are removed during topic representation through the vectorizer used to construct c-TF-IDF representations, following BERTopic's recommended approach to stopword removal (Grootendorst, 2022).

The patent-specific stopword procedure uses a 20,000-patent sample drawn uniformly from four technologically heterogeneous domains: aircraft, IoT, retail technologies, and robotics. The domains were chosen for their heterogeneity, not to represent the application corpus: terms that are frequent in all four are likely to reflect patent drafting conventions rather than technical content. The list combines (i) a baseline English stopword list, (ii) multi-domain intersection analysis for terms occurring frequently across all four domains, (iii) low TF-IDF terms in the bottom 15% of IDF with at least 20% document coverage, (iv) low BM25-IDF terms using the same quantile logic, (v) consensus between TF-IDF and BM25, and (vi) a patent-jargon layer for legal/formal terms such as 'embodiment', 'apparatus', 'comprising' and 'wherein'. The term sets obtained from components (i)–(iv) and (vi) are denoted by $V$ with the corresponding subscripts; the final stopword list $V$ used in Algorithm 1 and Figure 3 is:

$$V = V_{\text{baseline}} \cup V_{\text{multi-domain}} \cup (V_{\text{TF-IDF}} \cap V_{\text{BM25}}) \cup V_{\text{patent-jargon}} \tag{3}$$

This follows Sarica and Luo (2021), who show that general-purpose stopword lists miss uninformative terms in technical text. Figure 3 shows how the list is constructed.

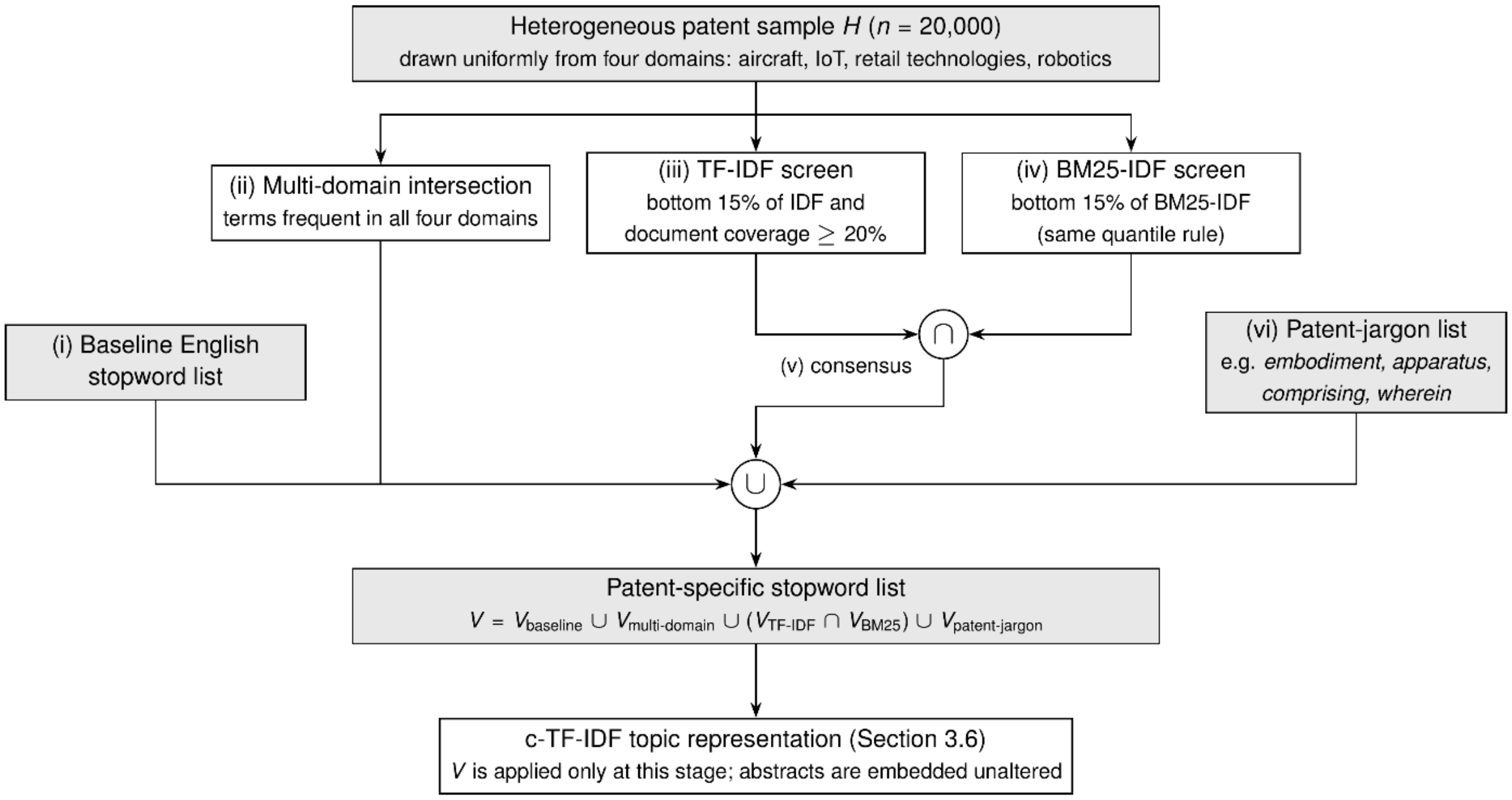

**Figure 3.** Patent-specific low-information term construction methodology.

## 3.5 Functional specialization of representation models

We use different representation models for document clustering and post-clustering topic comparison because the two tasks differ. PatentSBERTa is used to encode patent texts for clustering. Built on the sentence-transformer architecture (Reimers and Gurevych, 2019), it was fine-tuned on patent claims for patent distance and classification tasks (Bekamiri et al., 2024); we apply it to abstracts. The resulting 768-dimensional representations are passed to UMAP and HDBSCAN.

After topics have been formed, Qwen3-Embedding-0.6B (Zhang et al., 2025) is used to encode topic keyword representations for inter-topic similarity and visualization. Cosine similarity between the embedding vectors $\mathbf{a}$ and $\mathbf{b}$ of two topics is calculated as:

$$\text{sim}(\mathbf{a}, \mathbf{b}) = \frac{\mathbf{a} \cdot \mathbf{b}}{\|\mathbf{a}\| \|\mathbf{b}\|} \tag{4}$$

Different semantic tasks need not share one representation space. The clustering stage prioritizes patent-domain representation, while the refinement and visualization stages use a lightweight semantic representation of topics. We did not conduct an ablation comparing single- and dual-embedding approaches, so the use of two models is a design choice, not a demonstrated improvement.

### 3.6 Corpus-adaptive topic discovery and quality-guided refinement

Topic discovery follows the BERTopic architecture with patent-specific modifications. Full patent abstracts are embedded using PatentSBERTa without lemmatization. UMAP reduces the high-dimensional document vectors using cosine distance, and HDBSCAN forms density-based clusters with the excess-of-mass cluster selection method. Key settings are scaled to corpus size (Table 3), because the category corpora range from 591 to 29,215 patents.

Because HDBSCAN does not take the number of clusters as an input, we explore corpus-dependent parameter ranges and score the resulting solutions rather than search for an optimal number of topics. Each candidate solution is scored with a heuristic criterion that combines the document coverage $C_{c,\theta}$, rescaled silhouette $\mathrm{Sil}_{c,\theta}$ (Rousseeuw, 1987) and topic diversity $\mathrm{Div}_{c,\theta}$ (Dieng et al., 2020) of the solution obtained with configuration $\theta$ for corpus $D_c$:

$$Q_{c,\theta} = 0.40C_{c,\theta} + 0.30\mathrm{Sil}_{c,\theta} + 0.30\mathrm{Div}_{c,\theta} \quad (5)$$

The weights are heuristic. Coverage receives the largest weight to prioritize assigning documents to non-outlier topics, while silhouette and diversity balance geometric separation and distinctiveness of topic vocabularies. $Q_{c,\theta}$ is used to select configurations, not to validate them. Coverage and diversity are expressed in [0, 1], and the silhouette is rescaled to [0, 1] as (s + 1)/2 before weighting, where s is the raw silhouette. In Eq. (5), Sil denotes this rescaled value rather than the raw silhouette.

Documents identified as outliers are eligible for probability-based reassignment using the threshold $\tau_o = 0.05$. This is a soft-clustering membership threshold, not a calibrated probability that the assignment is correct. After initial clustering, small topics are refined hierarchically. Similar small topics are merged first; remaining small topics are then merged into larger topics when their semantic similarity exceeds the threshold $\tau_m$ = 0.35, measured as the cosine similarity (Eq. 4) between the Qwen3-Embedding vectors of the two topics' ten highest-ranked keywords. A topic is treated as small when it contains fewer than 100 patents; this bound is derived from 0.5–1% of the corpus size and clamped to the range 100–150. Small topics whose semantic similarity to every other topic is below 0.35 are dissolved and their patents are returned to the outlier set. The final c-TF-IDF representations are recomputed after refinement using the patent-specific stopword list. Figure 4 summarizes the pipeline.

**Table 3.** Corpus-size-dependent parameter ranges used in the implementation.

| Corpus size | Min. cluster size | Min. samples | UMAP neighbors | UMAP components | n-gram range | Min. df |
|---|---|---|---|---|---|---|
| <1,000 | 1.5% (min 15) | 3 | 20 | 6 | (1,2) | 2 |
| 1,000-2,000 | 1.2% (min 20) | 3 | 20-25 | 6 | (1,2) | 2 |
| 2,000-5,000 | 0.6% (min 25) | 3-5 | 25-30 | 6-8 | (1,2) | 3 |
| 5,000-10,000 | 0.6% (min 50) | 5 | 30-45 | 8 | (1,3) | 3-5 |
| >10,000 | 0.5% (min 70) | 5 | 45-50 | 8 | (1,3) | 5 |

### 3.7 Evidence-grounded LLM-assisted topic labeling

The final topic structure is converted into concise labels using GPT-5-mini (OpenAI, 2025a). For each topic, the input package contains the top 10 c-TF-IDF keywords and 10 representative patents. The prompt requests a concise topic label of 3–5 words, a short rationale, and a confidence category, all formatted according to a strict JSON schema.

Representative patents are included because keywords alone can be ambiguous or dominated by shared technical vocabulary. The patents constrain each label to the content of its cluster, and each label can be traced to both its keywords and its documents. The model's self-reported confidence is recorded but not used as evidence of label quality.

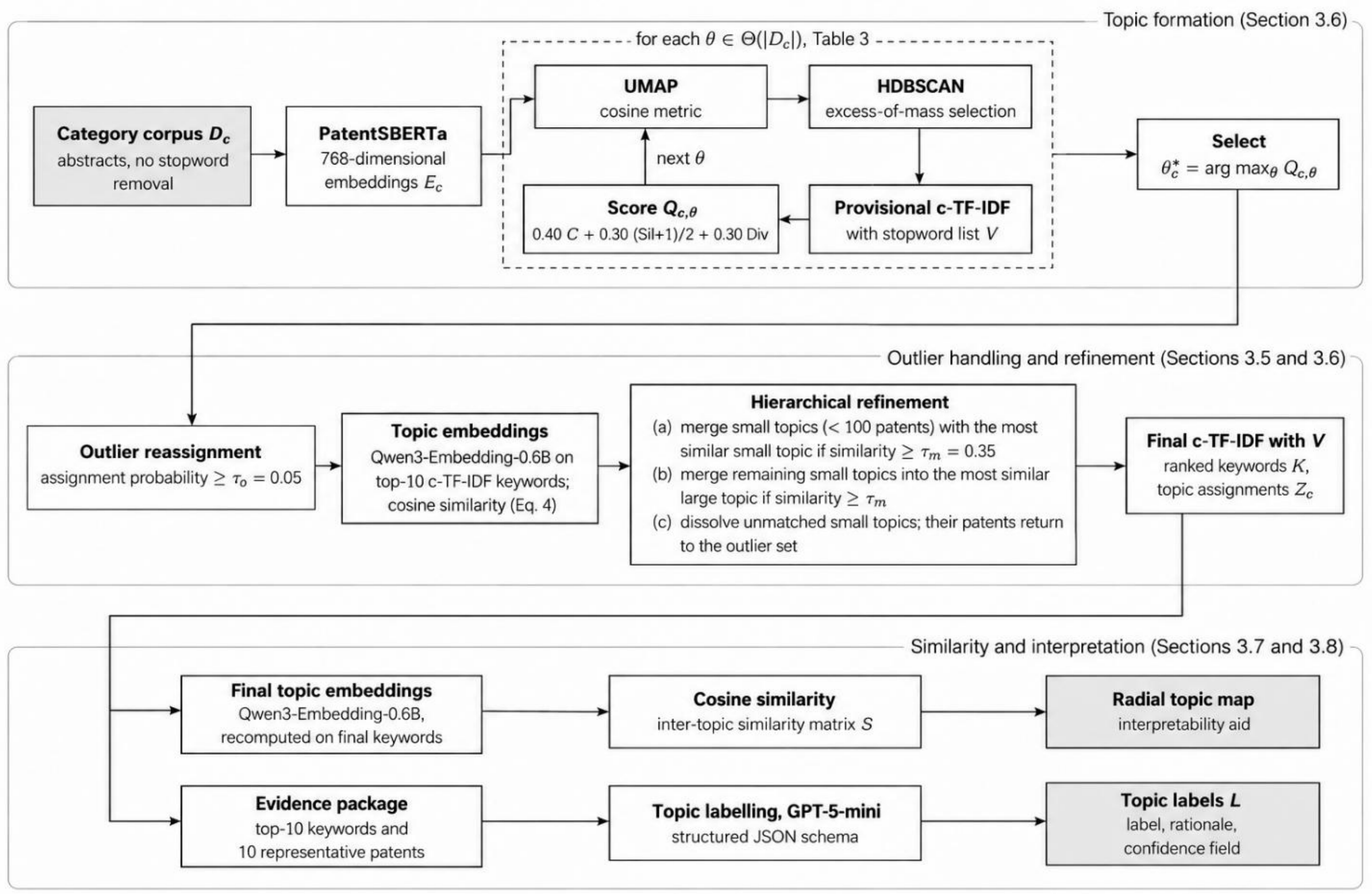

**Figure 4.** Corpus-adaptive topic discovery and refinement pipeline.

### 3.8 Visualization and structured exports

The framework produces an embedding-driven radial visualization and structured machine-readable exports. Topic similarities are calculated from Qwen3 topic embeddings. The largest topic is positioned at the center of the radial layout, and other topics are placed according to similarity to the central topic. High-similarity relationships can be displayed as connecting edges, while circle size represents topic document count. Within each topic, keywords are positioned according to c-TF-IDF importance and semantic relationships.

Structured exports include topic labels, confidence fields, document counts, ranked keywords with c-TF-IDF scores, representative patents, and pairwise topic similarity matrices. At the document level, consensus outputs preserve model votes and final labels. These exports allow auditing, manual inspection, and further analysis without rerunning the pipeline.

## 4. Demonstration Design

### 4.1 Demonstration corpus

The pipeline was applied to patent records retrieved from Lens.org. An optional three-LLM relevance filter was used to construct a domain-specific demonstration corpus, after which 30,964 patents remained. The same corpus was classified by novelty and innovation type. Topic models were then constructed separately for Radical, Incremental, Product, Process, and Service patent subsets. Because innovation type is multi-label, the Product, Process, and Service subsets are not mutually exclusive.

### 4.2 Demonstration criteria and scope of evidence

We assess the pipeline as a demonstration in the sense of design science research: the aim is to show that the pipeline can be applied to a realistic problem and produces the intended outputs (Hevner et al., 2004; Venable et al., 2016). Four criteria follow from the design principles: (i) output completion, that is, whether records receive structured outputs and unresolved votes are flagged; (ii) operation across corpus sizes, that is, whether a single procedure produces topic solutions across the reported size range; (iii) coverage and distinctiveness of the resulting topics; and (iv) traceability, that is, whether every label can be traced to model votes, keywords, and representative patents.

We report document coverage, topic diversity, final topic counts, structured-output success, and the labeling model's self-reported confidence. Coverage is the share of documents assigned to non-outlier topics. Topic diversity is the proportion of unique words among the top-k words of all topics (Dieng et al., 2020). The heuristic composite configuration score in Eq. (5) was used to select or refine candidate topic solutions. Because the score was used for tuning, reporting it as a validation result would be circular. Coverage and diversity are therefore presented primarily as descriptive indicators of the selected solutions. During development, the authors inspected a sample of topics and judged more than 90% of the generated labels to be consistent with the associated keywords and patents. Because this inspection did not follow a formal coding protocol, we report it only as indicative. Likewise, model consensus is not equated with classification accuracy: the demonstration documents execution and traceability within the selected corpus, while label accuracy remains untested.

### 4.3 Implementation logic and reproducibility

The three consensus models were invoked through an Ollama-based execution layer (Ollama, 2025), while GPT-5-mini was used for topic labeling through an API-based workflow orchestrated with LangChain (Chase, 2022). The classification system processed patent batches in parallel and used structured JSON parsing with cascade handling of malformed outputs. The method is specified at the level of model interfaces, so any environment that returns the required structured outputs can run it. Reproducing the results requires the exact model identifiers, complete prompt templates, category decision rules, structured-output schemas, inference settings, and deployment configurations, together with the model-vote records. These specifications are necessary because prompt wording, model revisions, and quantization choices can affect the resulting classifications. Table 4 lists the specifications for each component.

## 5. Results

### 5.1 Multi-LLM classification outputs

The consensus procedure produced novelty labels and binary innovation-type decisions for the 30,964-patent demonstration corpus. Table 5 reports the final category counts. Patents labeled Radical represented 5.65% of the corpus and Incremental patents 94.35%. Innovation type was evaluated independently: 93.93% of patents received a Product label, 11.05% a Process label, and 1.91% a Service label. Because type assignment is multi-label, category shares can sum to more than 100%; 2,111 patents (6.82%) received multiple type labels. These shares describe model-assigned labels within the selected corpus, rather than the prevalence of innovation types across the public sector.

**Table 4.** Reproducibility specification of the analytical components.

| Component | Input | Model / method | Key configuration | Output |
|---|---|---|---|---|
| Domain relevance filter | Title, abstract, metadata | Three distinct LLMs | 2-of-3 majority rule | Retain / exclude + model votes |
| Innovation classification | Patent evidence | Three distinct LLMs | Dimension-wise 2-of-3 voting | Novelty + multi-label type + votes |
| Document representation | Full patent abstract | PatentSBERTa | No stopword removal | 768-dimensional document vectors |
| Dimensionality reduction | Document vectors | UMAP | Corpus-size-scaled settings; cosine metric | Reduced vectors |
| Topic formation | Reduced vectors | HDBSCAN | Corpus-size-scaled density settings | Cluster assignments + outliers |
| Topic representation | Refined clusters | c-TF-IDF | Patent-specific stopwords applied here | Ranked topic keywords |
| Topic similarity | Topic keyword representations | Qwen3-Embedding-0.6B | Cosine similarity | Pairwise topic similarities |
| Topic labeling | Keywords + representative patents | GPT-5-mini | Structured JSON schema | Label + rationale + confidence field |

**Table 5.** Multi-LLM consensus classification outputs in the demonstration corpus.

| Output dimension | Category | Patents | Share (%) |
|---|---|---|---|
| Corpus | Total post-filtering | 30,964 | 100.00 |
| Novelty | Radical | 1,749 | 5.65 |
| Novelty | Incremental | 29,215 | 94.35 |
| Innovation type | Product | 29,086 | 93.93 |
| Innovation type | Process | 3,421 | 11.05 |
| Innovation type | Service | 591 | 1.91 |
| Innovation type | Multi-type overlaps | 2,111 | 6.82 |

## 5.2 Topic discovery across heterogeneous corpus sizes

The five category-specific corpora varied substantially in size, from 591 Service patents to 29,215 Incremental patents. After corpus-adaptive configuration and hierarchical refinement, the five separately modeled corpora produced 74 category-specific topic instances. Because the Product, Process, and Service corpora can overlap and the novelty dimension is modeled separately, these 74 instances should not be read as mutually exclusive topics in one global patent landscape. Table 6 summarizes the outcomes. Coverage ranged from 94.4% to 100%, and topic diversity ranged from 0.900 to 1.000. Macro-average coverage across the five category-specific models was 98.1%. The selected solutions thus retained most documents and produced distinct top-word sets at all corpus sizes.

The configuration scores in Table 6 are retained for transparency because they were part of the configuration-selection logic. They should not be interpreted as independent evidence of superiority over alternative topic models, and the macro-average values are descriptive summaries. Outlier reassignment can increase coverage, and merging changes the number of topics. The final indicators therefore do not isolate the effects of adaptive settings or the stopword list.

**Table 6.** Topic discovery outcomes across innovation categories.

| Category | Patents | Final topics | Coverage (%) | Topic diversity | Reported configuration score |
|---|---|---|---|---|---|
| Radical | 1,749 | 8 | 100.0 | 0.913 | 0.831 |
| Incremental | 29,215 | 24 | 94.4 | 0.925 | 0.811 |
| Product | 29,086 | 25 | 96.1 | 0.916 | 0.814 |
| Process | 3,421 | 9 | 99.9 | 0.900 | 0.829 |
| Service | 591 | 8 | 100.0 | 1.000 | 0.858 |
| Macro-average | - | - | 98.1 | 0.931 | 0.829 |

### 5.3 Structured output reliability and topic interpretation

More than 99% of patent classification outputs were parseable; fewer than 1% returned UNKNOWN and required manual attention. For topic labeling, all 74 category-specific topic instances received HIGH confidence in the model-generated confidence field, and the structured JSON parser achieved complete schema compliance for the topic-label outputs. The informal inspection described in Section 4.2 found more than 90% of sampled labels consistent with their keywords and patents. These figures indicate structured-output completion in this run. Uniform HIGH confidence does not distinguish reliable labels from unreliable ones, and neither parsing success nor the informal inspection provides an independent accuracy estimate.

### 5.4 Synthesis of methodological findings

**RQ1** asked how several LLMs can be combined to classify patents along two innovation dimensions while making each decision inspectable. Separating the two dimensions, voting on each type label independently and storing every model vote produced novelty labels and multi-label type assignments for all 30,964 patents; fewer than 1% of classification outputs returned UNKNOWN.

**RQ2** asked how one topic-modeling procedure can serve corpora of very different sizes. Scaling the UMAP, HDBSCAN and vectorizer settings to corpus size (Table 3) and selecting among candidate configurations with the composite score produced topic solutions for corpora of 591 to 29,215 patents, with 94.4–100% document coverage and topic diversity of 0.900–1.000. The procedure thus produces topic assignments across the reported range under corpus-dependent settings. Comparative topic quality and the benefit of adaptation over fixed-parameter alternatives remain untested.

**RQ3** asked how LLM-generated topic labels can remain traceable to their evidence. Supplying the labeling model with both ranked keywords and representative patents, and exporting labels, rationales, keywords, representative patents and similarity matrices as linked records, make every label traceable to the material from which it was derived. All 74 label outputs complied with the schema. Traceability enables review of a label; its semantic adequacy still requires assessment against the topic documents.

## 6. Discussion

### 6.1 Contribution in relation to prior work

The main contribution of this study lies in coordinating analytical roles within a single patent analytics pipeline. The pipeline draws on established methods (sentence transformers, PatentSBERTa, UMAP, HDBSCAN, c-TF-IDF, and LLM-based annotation). It specifies which task each method performs, at which stage information is preserved or filtered, and how the output of one stage becomes an auditable input to the next.

This design connects two strands of recent work. Studies that apply BERTopic to patents (Jeon et al., 2023; Maghsoudi et al., 2025) focus on identifying and tracking technological themes. LLM-based annotation of patent text (Pelaez et al., 2024) has relied on a single generative model to assign conceptual labels. The present pipeline connects conceptual assignments to topic structures by retaining records of decisions made between stages. Patents are first classified by novelty and innovation type through a multi-model vote, and the resulting categories then define the corpora in which topics are discovered and labeled. Because the interfaces between stages are explicit, individual components can be replaced, benchmarked, or recalibrated as better models become available while the overall analytical logic remains unchanged.

## 6.2 Design insights from the demonstration

*Consensus classification.* Separate model votes make agreement and disagreement observable, so LLM annotation becomes a reviewable step. Evaluating each innovation type independently preserved overlap that a single-class scheme would suppress: 2,111 patents (6.82%) received more than one type label. Panels of diverse models have been found to reduce the idiosyncrasies of single LLM judges (Verga et al., 2024). The stored votes here allow consensus, borderline, and contested cases to be identified for targeted review. A low parsing-failure rate does not reveal whether model votes are unanimous or contested, so agreement patterns remain a separate evaluation target.

*Representation-task separation.* Abstracts were embedded without preprocessing, and the patent-specific stopword list was applied only when clusters were translated into c-TF-IDF keywords. This follows the observation that general-purpose stopword lists miss uninformative terms in technical text (Sarica and Luo, 2021). The resulting topic descriptions showed limited repetition among their top words, with topic diversity of 0.900–1.000 across the five corpora. A separate, lightweight embedding model was used to compare topics. This places topic comparison in a separate representation space. It does not provide independent validation because the topic keywords originate from the same corpus.

*Corpus-sensitive topic formation.* A single procedure used UMAP, HDBSCAN, and vectorizer settings scaled to corpus size to produce topic solutions for corpora ranging from 591 to 29,215 patents, a factor of about 50, with document coverage of 94.4–100% under the stated size-dependent settings. Outlier reassignment and hierarchical refinement then converted raw cluster assignments into a refined set of topics.

*Evidence-grounded interpretation.* Supplying the labeling model with both ranked keywords and representative patents ties each label to the evidence from which it was derived. All 74 label outputs complied with the output schema. Because keywords, representative patents, labels and rationales are stored as linked records, an analyst can inspect or revise a label without reclustering.

## 6.3 Implications for research

For innovation measurement, LLM-based classification offers a text-based judgment of novelty and innovation type that complements citation-based and textual indicators (Dahlin and Behrens, 2005; Verhoeven et al., 2016; Arts et al., 2021; Kelly et al., 2021). Unlike forward-citation measures, it is available as soon as a patent is published. This makes it a candidate for screening recent technological activity, subject to validation of the novelty decisions. Comparing these judgments with citation-based indicators could show where the measures converge or diverge. For patent analytics methodology, the explicit stage interfaces and the audit record supply the material needed for component-wise evaluation. Future studies can therefore test alternative embedding models, clustering settings or labeling strategies within the same pipeline.

## 6.4 Implications for practice

Once patents have been classified and topics discovered, the structured outputs can be linked to filing dates, applicants, inventors, jurisdictions, citations, or patent families without changing the core text pipeline. Potential applications include technology opportunity discovery, emerging-theme monitoring, competitor mapping, roadmapping, and policy analysis. Technology scouts and R&D managers can screen large patent sets for candidate radical or service-oriented inventions and review the flagged cases alongside the model votes underlying them. Patent offices and policy analysts can use the labels to monitor how innovation types are distributed across technological themes, after checking a sample. Because each stage is modular, an organization can also replace a model, for example, with a locally hosted alternative, without redesigning the analysis.

# 7. Conclusion, limitations, and future research

This study presented a pipeline that converts large patent corpora into innovation classifications and labeled technological topics. It is organized around four design principles: consensus-based conceptual classification, representation-task separation, corpus-sensitive topic formation and refinement, and evidence-grounded interpretation. Applied to 30,964 public-sector-related patents, the pipeline classified the corpus by novelty and innovation type, with fewer than 1% of outputs requiring manual attention. It produced 74 labeled category-specific topic instances across category corpora containing 591 to 29,215 patents, with document coverage of 94.4–100%. These results suggest that assigning clear analytical roles to models can support corpus-scale execution and make analytical decisions reviewable.

The demonstration establishes feasibility and traceability; the accuracy of the pipeline remains to be established. The classification labels are model-based judgments, and errors shared across models can survive a majority vote (Kim et al., 2025). Future work should compare individual and consensus labels with expert coding of a stratified sample. It should also benchmark the topic layer against LDA, default BERTopic, and fixed-parameter configurations, and assess topic coherence, stability across random seeds, and label quality with human judges. These evaluations would also show what individual components contribute, such as the patent-specific stopword list, the corpus-scaled parameters, and the second embedding model. Validation should cover relevance filtering as well as innovation classification, because excluded records cannot enter the topic map and category errors change the documents clustered together.

The study is also limited by the scope of its data. The implementation uses English-language titles and abstracts, so information in claims and full descriptions is not used, and the LLMs may have encountered some of the patents during training. Applying the pipeline to other patent domains, languages, or technical document types would require recalibrating the stopword list, parameter ranges, and prompts. The source export's concentration in Chinese publications and utility models limits the basis for generalizing the demonstration to other patent systems. Document-level deduplication also allows family-related records to remain, potentially affecting category shares and topic sizes.

## Appendix A Patent search strategy and corpus characteristics

The Lens.org search was conducted on 26 April 2025 and returned 46,375 patent records. The search covered titles, abstracts and claims and was restricted to active records published from 1 January 2014 onwards. No upper publication date or restrictions on jurisdiction or document type were specified. The complete Boolean expression is reproduced below.

### Search expression

```
(title:("local government" OR "municipal government" OR "municipality" OR
"municipalities" OR "municipal" OR "city government" OR "city council" OR "city
councils" OR "town government" OR "town council" OR "town councils" OR "county
government" OR "county council" OR "county councils" OR "district government" OR
"district council" OR "district councils" OR "borough government" OR "borough
council" OR "borough councils" OR "village government" OR "village council" OR
"village councils" OR "parish government" OR "local authority" OR "local authorities"
OR "local administration" OR "regional government" OR "regional governments" OR
"community government" OR "community governments" OR "municipal services" OR
"municipal infrastructure" OR "local governance" OR "city management" OR "urban
administration" OR "local public works" OR "local utilities") OR abstract:("local
government" OR "municipal government" OR "municipality" OR "municipalities" OR
"municipal" OR "city government" OR "city council" OR "city councils" OR "town
government" OR "town council" OR "town councils" OR "county government" OR "county
council" OR "county councils" OR "district government" OR "district council" OR
"district councils" OR "borough government" OR "borough council" OR "borough
councils" OR "village government" OR "village council" OR "village councils" OR
"parish government" OR "local authority" OR "local authorities" OR "local
administration" OR "regional government" OR "regional governments" OR "community
government" OR "community governments" OR "municipal services" OR "municipal
infrastructure" OR "local governance" OR "city management" OR "urban administration"
OR "local public works" OR "local utilities") OR claim:("local government" OR
"municipal government" OR "municipality" OR "municipalities" OR "municipal" OR "city
government" OR "city council" OR "city councils" OR "town government" OR "town
council" OR "town councils" OR "county government" OR "county council" OR "county
councils" OR "district government" OR "district council" OR "district councils" OR
"borough government" OR "borough council" OR "borough councils" OR "village
government" OR "village council" OR "village councils" OR "parish government" OR
"local authority" OR "local authorities" OR "local administration" OR "regional
government" OR "regional governments" OR "community government" OR "community
governments" OR "municipal services" OR "municipal infrastructure" OR "local
governance" OR "city management" OR "urban administration" OR "local public works" OR
"local utilities"))
```

## Corpus construction and composition

Following deduplication by Lens ID, the initial corpus comprised 46,375 documents. The analytical procedure used English-language titles and abstracts, and public-sector relevance was assessed independently by Kimi K2, DeepSeek-V3.1, and gpt-oss-120b. Retention required at least two positive judgments. The final analytical corpus comprised 30,964 documents (66.8%), a net reduction of 15,411 records from the initial corpus. The initial corpus included 695 records (1.5%) without abstracts.

Table A1. Composition of the initial corpus before relevance filtering

| Dimension | Category | Documents | Percent |
|---|---|---|---|
| **Jurisdiction** | China | 41,107 | 88.6 |
| | United States | 3,181 | 6.9 |
| | European Patent Office | 647 | 1.4 |
| | Other jurisdictions | 1,440 | 3.1 |
| **Document type** | Chinese utility models | 34,809 | 75.1 |
| | Patent applications | 8,310 | 17.9 |
| | Granted patents | 3,181 | 6.9 |
| | Other documents | 75 | 0.2 |

Note. N = 46,375. Percentages are calculated within each dimension and may not sum to 100 because of rounding. Chinese utility models were reported under the Lens category "Limited Patent".

## Analytical scope

The retrieval strategy identifies disclosures containing terms associated with local government and municipal functions. Relevance screening is necessary because keyword matches can also reflect unrelated uses of these expressions. The resulting corpus represents a selected set of patent documents with potential public-sector applications. The active-status restriction and reliance on English-language text limit its coverage. Lens-ID deduplication operates at the document level; document counts should therefore be distinguished from counts of unique inventions. The jurisdiction and document-type distributions in Table A1 describe the initial corpus and should not be assumed to characterize the retained corpus.